\documentclass{article}

\usepackage{natbib}
\usepackage{doi}

\usepackage[T1]{fontenc}
\usepackage{amsmath}
\usepackage{amssymb}
\usepackage{svg}
\usepackage{booktabs} 
\usepackage{graphicx,verbatim}
\usepackage{url} 
\usepackage{arxiv}

\title{Performance vs Consistency: Evaluating a Foundation Model in Lung‑RADS Screening}

\author{
Benjamin Renoust\textsuperscript{ * 0009-0003-4047-0269}\\ 
Median Technologies, eyonis\textsuperscript{\textregistered}, Valbonne, France \\
The University of Osaka, D3 Center, Japan
\And 
{\textbf{Pierre Baudot\textsuperscript{ * 0000-0002-5574-6809}}}\\
Median Technologies, eyonis\textsuperscript{\textregistered}, Valbonne, France  
\And Tiffany Foriel\textsuperscript{ 0009-0005-7651-925X}\\
Median Technologies, eyonis\textsuperscript{\textregistered}, Valbonne, France   
\And Yousra Haddou\textsuperscript{ 0009-0009-5592-3818} 
\\
Median Technologies, eyonis\textsuperscript{\textregistered}, Valbonne, France  
\And Charles Voyton\textsuperscript{ 0009-0003-4047
-0269}\\
Median Technologies, eyonis\textsuperscript{\textregistered}, Valbonne, France  
\And Pierre-Henri Siot\textsuperscript{ 0009-0006-7098-8300}\\
Median Technologies, eyonis\textsuperscript{\textregistered}, Valbonne, France  
\And Ezequiel Geremia\textsuperscript{ 0009-0008-7328-6720}\\
Median Technologies, eyonis\textsuperscript{\textregistered}, Valbonne, France  
\And Danny Francis\textsuperscript{ 0009-0009-3185-0693}\\
Median Technologies, eyonis\textsuperscript{\textregistered}, Valbonne, France  
\And Jean-Christophe Brisset\textsuperscript{ 0000-0002-7947-3622}\\
Median Technologies, eyonis\textsuperscript{\textregistered}, Valbonne, France  
\And Valérie Bourdès\textsuperscript{ 0009-0003-1329-0684} \\
Median Technologies, eyonis\textsuperscript{\textregistered}, Valbonne, France  
\And Sylvain Bodard\textsuperscript{ 0000-0001-5381-3361}\\
Median Technologies, eyonis\textsuperscript{\textregistered}, Valbonne, France \\
Université de Paris Cité, AP-HP, Hôpital Universitaire Necker Enfants Malades, \\
Service d’Imagerie Adulte, Paris, France. \\
Memorial Sloan Kettering Cancer Center, Department of Radiology, NY, USA. \\
Massachusetts General Hospital, Center for Transplantation Sciences, Harvard Medical Shool, Boston, USA. \\
Sorbonne Université, CNRS UMR7371, INSERM U1146, LIB, Paris, France.
\And Benoit Huet\textsuperscript{ 0000-0002-0608-6939}
\\
Median Technologies, eyonis\textsuperscript{\textregistered}, Valbonne, France  
\And{\textsuperscript{* }These authors contributed equally to this work.}\\
\small \texttt{firstname.lastname@mediantechnologies.com}
}

\renewcommand{\shorttitle}{Performance vs Consistency: Foundation Model vs Lung-RADS}
\date{August 10, 2026}

\begin{document}
  
\maketitle              
\begin{abstract}
Foundation models have recently demonstrated strong capabilities across a wide range of medical imaging tasks. However, their performance in structured clinical interpretation settings remains insufficiently explored. In lung cancer screening, interpretative variability persists despite standardized frameworks such as Lung-RADS.
In this study, we evaluate MedGemma, a medical general-purpose foundation model derived from Gemini and its fine-tuned version adapted for lung cancer detection and diagnosis, compared against radiologists performing Lung-RADS\textsuperscript{\textregistered} v2022 assessment on the NLST dataset. Twelve radiologists independently evaluated each case in a multi-reader design, enabling quantification of inter-reader variability.
Radiologists achieved a mean AUC of 0.90, with substantial variability across readers (range: 0.80–0.94). The native foundation model achieved an AUC of 0.70, failing to reach clinically relevant performance. In contrast, fine-tuning significantly improved performance to an AUC of 0.83, placing the model within the lower range of individual radiologists performance. 
These findings highlight a trade-off between peak accuracy and prediction consistency. 
Unlike radiologists, under fixed conditions, the model produces deterministic outputs, removing inter-run variability under identical inputs, in contrast to inter-reader variability observed among radiologists.
This supports the role of fine-tuned foundation models 
potential complementary tools for clinical decision support, particularly in settings with limited expertise. 
However, evaluation is performed on a case-enriched cohort from NLST and does not account for real-world prevalence or external validation, limiting direct clinical generalization.

\end{abstract}

\section{Introduction}


Low-dose computed tomography (LDCT) screening has significantly improved early detection of lung cancer and reduced mortality in high-risk populations~\cite{national_lung_screening_trial_research_team_reduced_2011,de_koning_reduced_2020}. Structured reporting systems like Lung-RADS standardize interpretation and reduce false-positive rates by linking imaging findings to risk levels and management recommendations~\cite{chelala_lung-rads_2021,christensen_acr_2024-1}. Despite these efforts, inter-reader variability remains a key limitation, particularly for indeterminate nodules~\cite{nair_variable_2018,han_influence_variability2018}. Differences in nodule characterization lead to inconsistent categorization, directly impacting follow-up decisions.

Recent advances in artificial intelligence have enabled strong performance in lung cancer detection and risk prediction. Deep learning models trained on large datasets such as NLST, including end-to-end 3D convolutional approaches~\cite{ardila_end--end_2019,liao_evaluate_2019,bodard_rethinking_2025} and temporal risk models such as Sybil~\cite{mikhael_sybil_2023}, have demonstrated high diagnostic accuracy. More recently, foundation models trained on large-scale multimodal data have emerged as general-purpose systems capable of addressing a wide range of medical imaging tasks~\cite{sellergren_medgemma_2025,saab_capabilities_2024,wang_medclip_2022}. Vision-language architectures such as MedGemma extend this paradigm to clinical imaging by combining image and text representations.
However, the evaluation of foundation models in structured screening frameworks such as Lung-RADS remains limited, particularly when compared to multiple independent radiologist assessments. 

In this work, we investigate the following questions: (i) can a general foundation model perform malignancy prediction in a Lung-RADS setting, (ii) to what extent fine-tuning improves performance, and (iii) how model predictions compare to the variability observed across radiologists.
We evaluate MedGemma, a foundation model, in both native and task-specific adaptation (fine-tuned), against radiologists performing Lung-RADS\textsuperscript{\textregistered} v2022 assessment. Our contribution is not on providing a novel tuning approach, but instead focuses on comparing diagnostic performance while contextualizing the results with respect to inter-reader variability, providing a more comprehensive perspective on the role of AI in lung cancer screening.

\section{Related Work}

Lung-RADS\textsuperscript{\textregistered} is a structured decision framework that standardizes lung cancer screening interpretation through categorical risk levels linked to management recommendations. It has been shown to reduce false positive rates~\cite{chelala_lung-rads_2021,christensen_acr_2024-1}. 
However, inter-reader variability in lung cancer screening persists even with standardized frameworks. Studies report only moderate agreement in nodule characterization and Lung-RADS categorization, particularly for indeterminate cases~\cite{nair_variable_2018,han_influence_variability2018}. 

Deep learning models have demonstrated strong performance in lung cancer detection and malignancy prediction, particularly on large datasets such as NLST. End-to-end models, including those proposed by Ardila \textit{et al.}~\cite{ardila_end--end_2019}, Liao \textit{et al.}~\cite{liao_evaluate_2019}, Sybil~\cite{mikhael_sybil_2023}, and recently LCS~\cite{bodard_rethinking_2025} have achieved high diagnostic accuracy in screening populations.
More recently, large multimodal foundation models have been proposed as general-purpose systems capable of adapting to multiple medical imaging tasks~\cite{sellergren_medgemma_2025,hamamci2026generalist}. 
Their application to structured decision frameworks such as Lung-RADS remains only partially characterized.
MedGemma~\cite{sellergren_medgemma_2025} is currently, to our knowledge, the only version designed to process volumetric CT data as sequences of consecutive axial slices. By benchmarking naive and fine-tuned variants against multiple independent Lung-RADS readings, we evaluate performance in the context of inter-reader variability.

\section{Dataset and Study Design}

\subsection{Dataset Construction}

This study is conducted on a subset of chest CT scans derived from the National Lung Screening Trial (NLST), a large-scale multi-center lung cancer screening program with longitudinal follow-up and histopathology-confirmed outcomes.

The dataset comprises 481 patients, with 132 cancer cases (27.4\%) and 349 non-cancer cases (72.6\%). Cancer cases are defined by biopsy-confirmed diagnosis within one year following the CT scan, while non-cancer cases are confirmed by at least three years of follow-up without malignancy. We selected patients from the subset of fully annotated patients~\cite{Renoust_BAUDOT_etal_2026}.
The cohort is case-enriched (27.4\% cancer prevalence) to ensure statistical power for performance estimation; however, this design deviates from real-world screening prevalence and may affect metrics such as predictive values and decision thresholds.

In NLST, each patient may have multiple screening time points with several CT series. For this study, a single clinically relevant scan was selected per patient.  
For cancer cases, the selected scan corresponds to the acquisition closest to the time of diagnosis or within one year before diagnosis when the diagnostic scan was not available. For non-cancer cases, the earliest available screening scan was selected to maximize the duration of negative follow-up.
Scans with slice thickness greater than or equal to 5 mm, incomplete lung coverage, or prone acquisition were excluded, leaving 481 patients.
When multiple reconstruction series were available, a standardized selection protocol was applied to retain the most diagnostically relevant scan. Priority was given to thin-slice acquisitions (closest to 1 mm), and sharp lung kernels.
Scans were sourced from multiple manufacturers, reflecting real-world variability. 
Importantly, performance metrics reported here (AUC, sensitivity, specificity) remain prevalence-invariant, but operating points derived from the Maximum Youden Index may not directly translate to clinical screening settings.

\subsection{Study Design}

A multi-reader, multi-case (MRMC) inspired design was used to measure radiologist performance and inter-reader variability in Lung-RADS\textsuperscript{\textregistered} v2022 assessment.

\textit{Readers.}
Twelve board-certified radiologists participated in the study, including both junior ($<3$ years) and senior ($\geq3$ years of practice after certification) radiologists, enabling analysis of performance variability across experience groups. All readers were blinded to ground truth labels, clinical data, and the assessments of other readers.

\textit{Reading Protocol.}
Case allocation was performed such that each patient was assessed independently by three different readers, while maintaining a balanced distribution of cancer and non-cancer cases across readers. Each radiologist evaluated approximately 120 cases on average, depending on allocation within the MRMC design, cumulating 1,443 independent readings.
All readings were performed under a single time-point scenario, where no prior or follow-up imaging was available. Radiologists assigned a Lung-RADS\textsuperscript{\textregistered} v2022 category based solely on the CT scan, 
reflecting a baseline lung cancer screening setting.

Overall, the combination of controlled scan selection, case-enriched prevalence, and repeated independent readings provides a robust framework for analyzing both diagnostic accuracy and variability in lung cancer screening.

\subsection{Inter-Reader Variability}

Inter-reader variability is quantified using the ordinal mapping of the LungRADS score. We report in Table \ref{tab:var} evaluation of:
\begin{itemize}
\item Mean pairwise L1 (Manhattan) distance, to measure the flat distance between assessments so that close answers (i.e. by 1 score grading) are less penalized~\cite{krippendorff_content_2004}.
\item Fleiss' Kappa~\cite{landis_measurement_1977}, to measure the actual agreement as compared to random agreement.
\item Intraclass Correlation Coefficient (ICC)~\cite{mcgraw_forming_1996}, to measure the consistency across rating levels.
\end{itemize}

\begin{table}[t!]
\centering
\caption{Summary of reader variability measurements}
\begin{tabular}{lccc}
\toprule
\textbf{Method} & \textbf{All} & \textbf{Junior} & \textbf{Senior} \\
\midrule
L1 distance & 0.892 & 1.056 & 0.817 \\
Fleiss' Kappa & 0.329 & 0.235 & 0.379 \\
ICC & 0.683 & 0.580 & 0.738 \\
\bottomrule
\end{tabular}
\label{tab:var}
\end{table}

These results indicate reduced variability and higher consistency among senior readers compared to junior readers. The average L1 distance remains below one category difference.
Fleiss’ Kappa values fall within the range classified as fair agreement (0.21–0.40)~\cite{landis_measurement_1977}, consistent with prior studies reporting moderate inter-reader agreement in lung nodule and Lung-RADS assessment~\cite{nair_variable_2018,han_influence_variability2018}.
ICC values fall within the range typically interpreted as moderate reliability (0.50-0.75) according to established guidelines~\cite{koo_guideline_2016}.
Notably, both Kappa and ICC analyses show that junior readers tend to lie at the lower bound of the respective agreement and reliability ranges, whereas senior readers consistently approach the upper bound.

\section{MedGemma}

We evaluate MedGemma, a large-scale foundation model for medical imaging that produces a continuous malignancy score at the patient level. Two configurations are considered: (i) a native zero-shot version, and (ii) a task-specific fine-tuned version for Lung-RADS estimation. For more precise methodological details, our approach and experiments are made publicly available\footnote{\url{https://github.com/EYONIS-AIDS-DS/medgemma-miccai}}.

\textit{Input.} Each CT volume is convoluted through the z-axis. One convolution is a sequence of 25 consecutive axial slices, with a stride of 5, on which we apply the following strategy. The patient prediction is the maximum prediction over all convolutions. This max-aggregation strategy prioritizes sensitivity to focal malignant patterns.

\textit{Coarse-to-Fine prompting strategy.}
To adapt MedGemma to Lung-RADS assessment, we employ a coarse-to-fine (C2F) chain of thoughts prompting strategy that structures inference across multiple reasoning stages. The model first performs a global analysis of lung patterns, and identifies potential abnormalities across the selected slices. The model is asked to focus and report on the presence and features of potential nodules, refining its attention toward cancer suspicion areas. Finally, it produces a Lung-RADS-like assessment through logit extraction from token probability of each LungRADS class. This approach mimics radiologist reasoning from global inspection to targeted lesion analysis, and was found critical for achieving stable performance.

\textit{MedGemma (native).}
We evaluate the \texttt{google/medgemma-1.5-4b-it} vision-language model~\cite{sellergren_medgemma_2025,saab_capabilities_2024} in a zero-shot setting. 
Whole-scan prediction is obtained as the maximum score across all the slice sequences. The C2F design and slice selection were empirically optimized for stability.

\textit{MedGemma (fine-tuned).}
The same base model is further adapted to lung cancer screening using the NLST Train3 split from \cite{bodard_rethinking_2025}, which consists of 7,585 patients from NLST (393 cancer), with 21,533 manually annotated nodules localization, segmentation, and characterization (405 malignant). Fine-tuning is performed using Quantized Low-Rank Adaptation (QLoRA), with 4-bit NF4 quantization, rank 16, and scaling factor (alpha) of 32, applied to all linear layers of the model. Training used positive GT labels only for slice chunks containing malignant nodules, while all others are associated with negative (benign) labels, and a selective loss masking strategy in which only the single classification token drove the objective, using focal loss ($\gamma = 2.0$) to compensate for chunks imbalance, and balanced 1:1 sampling of positive and negative scans. Such training strategy greatly improved the original scan level and cross-entropy strategy. 

\textit{Output Representation.}
Both configurations produce malignancy scores on a shared continuous scale from 1 to 5 corresponding to LungRADS 1 to 4X with 4B and 4X merged into a single class 5 since 4X is only an aggravation feature and not a class on itself. This enables direct comparison with radiologist assessments, where LungRADS values are only added of 0 (LungRADS 0, when radiologists are unable to answer).


\begin{figure}[b!]
    \centering
    \includegraphics[width=0.6\textwidth]{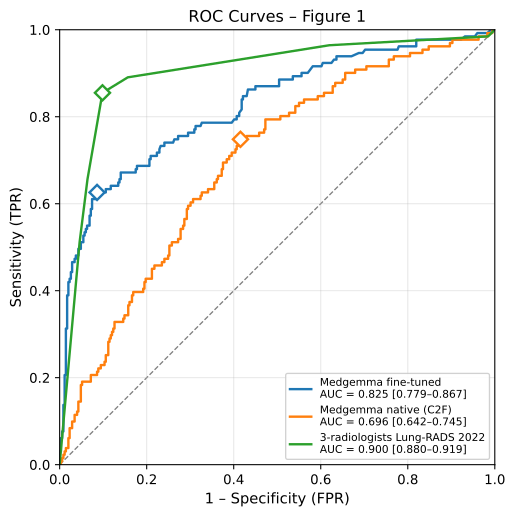}
    \caption{Comparison of ROC performances between the 12 radiologists (Lung-RADS\textsuperscript{\textregistered} v2022), against both MedGemma C2F and MedGemma fine-tuned.}
    \label{fig:globalroc}
\end{figure}

\section{Evaluation Method}

Using the continuous ordinal scores of LungRADS as presented previously, diagnostic performance is evaluated with Area Under the ROC Curve (AUC), Sensitivity, Specificity, and Accuracy.
Ground truth is defined using NLST outcomes, based on biopsy-confirmed diagnosis or long-term follow-up without cancer.
Sensitivity, specificity, and accuracy are computed at the optimal operating point defined by the Maximum Youden Index~\cite{steyerberg_clinical_2019} (MYI).

\subsection{Statistical Analysis}

Performance metrics are computed across all readers and cases, and Confidence Intervals estimated over 5,000 bootstraps samples. 
Since each radiologist is assigned a unique patient subset, AUC comparisons are performed using a one-sided Welch $t$-test over bootstrap samples.
Variability is assessed at the population level and stratified by reader experience.

\section{Results}

\subsection{Diagnostic Performance}

\begin{figure}[b!]
    \centering
    \includegraphics[width=1\textwidth]{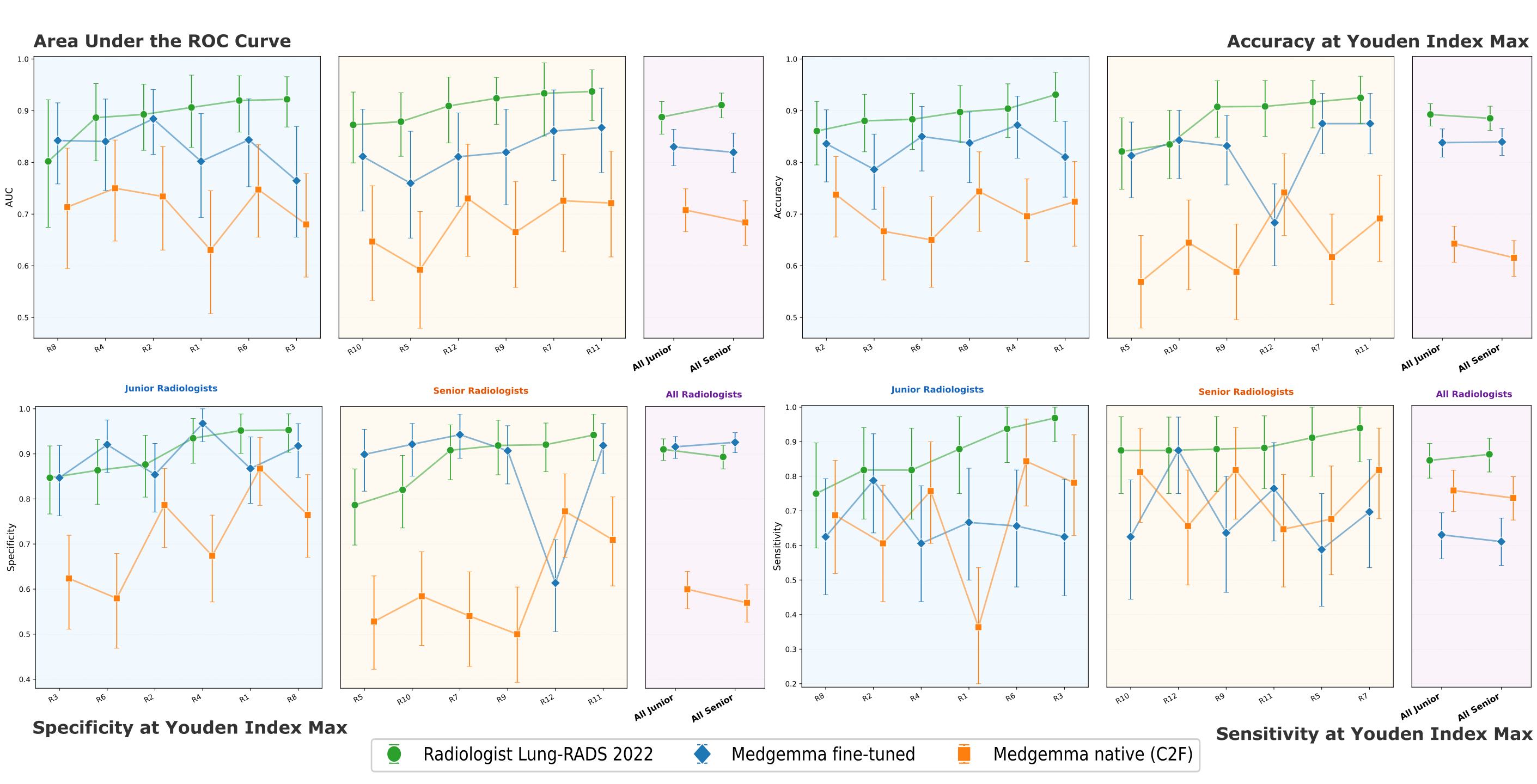}
    \caption{Comparative evaluation metrics in increasing radiologist performance order, and grouped by seniority: top left, AUC; top right, accuracy; bottom left, specificity; bottom right sensitivity. Since each radiologist is assigned a unique set, individual model performances are computed on the same set.}
    \label{fig:globalmetrics}
\end{figure}

When measuring AUC performance, although radiologists remain largely superior, we can observe that fine-tuning clearly improves the overall performance against C2F (see Figure~\ref{fig:globalroc}).
The native MedGemma model (C2F) achieved an AUC of 0.70 substantially lower than radiologist performance. In contrast, the fine-tuned MedGemma model achieved an AUC of 0.83, placing it approaching the lower bound of individual radiologist performance.
Radiologists achieved a mean AUC of 0.90, with observed variability across readers (see Figure~\ref{fig:globalmetrics}), indeed, 
the fine-tuned model matched or exceeded the performance of a subset of junior radiologists.
The best-performing reader achieved an AUC of 0.94, while the lowest-performing reader achieved 0.80 -- exact figures reported in Table~\ref{tab:metrics}.

This improvement is primarily driven by higher specificity, indicating that the fine-tuned model reduces false positive predictions compared to its native counterpart. 
The case-enriched design makes accuracy a more informative metric than in typical screening settings. 
We observe performances of the fine-tuned MedGemma (with 0.84) closing in on radiologist performance (against 0.89), and reaching the level of the least accurate radiologist. 
Although still below radiologist accuracy, the native MedGemma (C2F) shows superior sensitivity in comparison to the fine-tuned model, which can be explained by the position of the MYI along the ROC curve.

\begin{table}[b!]
\centering

\caption{Diagnostic performance comparison: comparative performance of the min radiologist is done on the exact same set.}
    \makebox[\linewidth][c]{
\begin{tabular}{lcccc}
\toprule
Method & AUC & Sensitivity & Specificity & Accuracy \\
\midrule
Radiologists (all) & \textbf{0.90 [0.88-0.92]} & \textbf{0.85 [0.82-0.89]} & 0.90 [0.88-0.92] &  \textbf{0.89 [0.87-0.91]}\\
MedGemma (C2F) & 0.70 [0.64-0.75] & 0.75 [0.67-0.82] & 0.58 [0.53-0.64] & 0.63 [0.59-0.67] \\
MedGemma (fine-tuned) & 0.83 [0.78-0.87] & 0.63 [0.54-0.71] & \textbf{0.91 [0.88-0.94]}
 & 0.84 [0.80-0.87] \\
 \midrule
Radiologists (min) & 0.80 [0.67-0.92] & \textbf{0.75 [0.59-0.90]} & 0.79 [0.70-0.87] & \textbf{0.82 [0.75-0.89]} \\
MedGemma (C2F) & 0.71 [0.59-0.83] & 0.69 [0.52-0.85] & 0.53 [0.42-0.63] & 0.57 [0.48-0.66] \\
MedGemma (fine-tuned) & \textbf{0.84 [0.76-0.92]}& 0.63 [0.46-0.79] & \textbf{0.90 [0.82-0.95]} & 0.81 [0.73-0.88]  \\
\bottomrule
\end{tabular}
}
\label{tab:metrics}
\end{table}

In terms of seniority performance comparison, we confirm the overall positive impact of diagnostics from senior readers on AUC, although junior readers tend to match or outperform seniors on accuracy and specificity. Interestingly, it is on these metrics where we observe the fine-tuned MedGemma closing the gap.

These results highlight a trade-off between sensitivity and specificity depending on model fine-tuning, emphasizing the importance of evaluating models across the full ROC curve rather than at a single operating point.

\section{Discussion}

This study provides a direct comparison between radiologists' Lung-RADS assessments and a foundation model under identical evaluation conditions.


The poor performance of the native foundation model indicates that general-purpose representations are not sufficient for this task, and that task-specific adaptation is required to achieve clinically relevant performance. Although the native model shows higher sensitivity at the MYI operating point, this reflects a different threshold selection along the ROC curve and a tendency to over-predict positive cases, rather than superior discriminative ability. In contrast, the higher AUC of the fine-tuned model demonstrates improved separation between malignant and non-malignant cases, indicating that the observed sensitivity difference arises from thresholding effects. Fine-tuning enables the model to reach performance levels comparable to the lowest tier of the less experienced radiologists.

In contrast to the variability observed among radiologists despite the use of Lung-RADS, the models produce near deterministic outputs (under identical prompt and context without history, temperature is nul). From a clinical standpoint, this property can help standardize decision-making and improve consistency, especially in settings with limited expertise or radiologist shortages.

While interesting, our results must be mitigated by the representativeness of our sample. Indeed, we are comparing against NLST patients only (the dataset dating from the early 2000's might not reflect the current practice). Our evaluation subset is also heavily balanced towards cancer patients (roughly 27\% against the observed 4\% from the NLST dataset). The experiment is a first step towards assessing AI performances in a real setting, but both training and evaluation are made on NLST. This means that we cannot exclude a risk of dataset overfitting from the FT version of MedGemma. Further ablation studies are necessary to help validate the importance of each step in our C2F refinement process and efficiency of our fine-tuning. To compensate, we have opened our code repository. Finally, deterministic outputs under identical conditions only demonstrate repeatability, further experiments such as repeated runs with prompt perturbations are required to assess robustness or clinical reliability.

\section{Conclusion and Future Work}

We assessed a foundation model and its fine-tuned version against radiologists' Lung-RADS assessment in lung cancer screening. While radiologists achieve higher peak performance, substantial variability exists.
Fine-tuned foundation models reach performance comparable to less experienced readers while providing fully consistent predictions. These findings emphasize the importance of considering both accuracy and reproducibility in the evaluation of AI systems for clinical use.
This study has several limitations. First, evaluation is restricted to a case-enriched subset of NLST without external validation, limiting generalizability.
Second, the deterministic nature of model inference was not evaluated under input perturbations or uncertainty estimation. Third, only a single aggregation strategy and prompting design were considered without ablation. 
Future work will measure inter-run variability, while extending the range of models, and evaluate the integration of models within radiologists' workflows.


\bibliographystyle{unsrtnat}
\bibliography{MEDIAN_minimal}

@misc{Renoust_BAUDOT_etal_2026, 
    title={LIDC-IDRI-like Multi-Reader annotations of lung nodules on NLST CT scans (LIDC-annot-NLST501)}, 
    url={https://www.cancerimagingarchive.net/analysis-result/lidc-annot-nlst501}, 
    DOI={10.7937/Y2DX-E334}, 
    publisher={The Cancer Imaging Archive}, 
    author={Renoust, Benjamin and Baudot, Pierre and Voyton, Charles M. and Chikbouni, dorian and Foriel, Tiffany and Haddou, Yousra and Geremia, Ezequiel and De Bie, Gwendoline and Le, Van Khoa and Bobin, Vincent and Francis, Danny and Siot, Pierre-Henri and Bodard, Sylvain and Bourdes, Valerie and Huet, Benoit}, 
    year={2026} 
}

@article{ardila_end--end_2019,
	title = {End-to-end lung cancer screening with three-dimensional deep learning on low-dose chest computed tomography},
	volume = {25},
	issn = {1546-170X},
	doi = {10.1038/s41591-019-0447-x},
	number = {6},
	journal = {Nature Medicine},
	author = {Ardila, Diego and Kiraly, Atilla P. and Bharadwaj, Sujeeth and Choi, Bokyung and Reicher, Joshua J. and Peng, Lily and Tse, Daniel and Etemadi, Mozziyar and Ye, Wenxing and Corrado, Greg and Naidich, David P. and Shetty, Shravya},
	year = {2019},
	pages = {954--961},
}

@inproceedings{bodard_rethinking_2025,
	address = {Chicago, IL, US},
	title = {Rethinking {Lung} {Cancer} {Screening}: {Longitudinal} {AI}/{ML} {Diagnostics} {Beyond} {Nodule} {Size} {Growth}},
	author = {Bodard, S. and Baudot, P. and Renoust, B. and Voyton, C. and De Bie, G. and Geremia, E. and Le, V.K. and Francis, D. and Siot, P.-H. and Haddou, Y. and Bourdès, V. and Huet, B.},
	year = {2025},
}

@article{chelala_lung-rads_2021,
	title = {Lung-{RADS} {Version} 1.1: {Challenges} and a {Look} {Ahead}, {From} the {AJR} {Special} {Series} on {Radiology} {Reporting} and {Data} {Systems}},
	volume = {216},
	issn = {1546-3141},
	doi = {10.2214/AJR.20.24807},
	number = {6},
	journal = {AJR. American journal of roentgenology},
	author = {Chelala, Lydia and Hossain, Rydhwana and Kazerooni, Ella A. and Christensen, Jared D. and Dyer, Debra S. and White, Charles S.},
	year = {2021},
	pages = {1411--1422},
}

@article{christensen_acr_2024-1,
	title = {{ACR} {Lung}-{RADS} v2022: {Assessment} {Categories} and {Management} {Recommendations}},
	volume = {21},
	issn = {1546-1440, 1558-349X},
	doi = {10.1016/j.jacr.2023.09.009},
	number = {3},
	journal = {Journal of the American College of Radiology},
	publisher = {Elsevier},
	author = {Christensen, Jared and Prosper, Ashley Elizabeth and Wu, Carol C. and Chung, Jonathan and Lee, Elizabeth and Elicker, Brett and Hunsaker, Andetta R. and Petranovic, Milena and Sandler, Kim L. and Stiles, Brendon and Mazzone, Peter and Yankelevitz, David and Aberle, Denise and Chiles, Caroline and Kazerooni, Ella},
	year = {2024},
	pages = {473--488},
}

@article{de_koning_reduced_2020,
	title = {Reduced {Lung}-{Cancer} {Mortality} with {Volume} {CT} {Screening} in a {Randomized} {Trial}},
	volume = {382},
	issn = {1533-4406},
	doi = {10.1056/NEJMoa1911793},
	number = {6},
	journal = {The New England Journal of Medicine},
	author = {de Koning, Harry J. and van der Aalst, Carlijn M. and de Jong, Pim A. and Scholten, Ernst T. and Nackaerts, Kristiaan and Heuvelmans, Marjolein A. and Lammers, Jan-Willem J. and Weenink, Carla and Yousaf-Khan, Uraujh and Horeweg, Nanda and van 't Westeinde, Susan and Prokop, Mathias and Mali, Willem P. and Mohamed Hoesein, Firdaus A. A. and van Ooijen, Peter M. A. and Aerts, Joachim G. J. V. and den Bakker, Michael A. and Thunnissen, Erik and Verschakelen, Johny and Vliegenthart, Rozemarijn and Walter, Joan E. and Ten Haaf, Kevin and Groen, Harry J. M. and Oudkerk, Matthijs},
	year = {2020},
	pages = {503--513},
}

@article{hamamci2026generalist,
  title={Generalist foundation models from a multimodal dataset for 3D computed tomography},
  author={Hamamci, Ibrahim Ethem and Er, Sezgin and Wang, Chenyu and Almas, Furkan and Simsek, Ayse Gulnihan and Esirgun, Sevval Nil and Dogan, Irem and Durugol, Omer Faruk and Hou, Benjamin and Shit, Suprosanna and others},
  journal={Nature Biomedical Engineering},
  pages={1--19},
  year={2026},
  publisher={Nature Publishing Group UK London},
}

@article{han_influence_variability2018,
  title={Influence of Observer Variability on Lung Nodule Classification and Lung-RADS Categorization},
  author={Han, Dong and Heuvelmans, Marjolein A. and Oudkerk, Matthijs},
  journal={European Radiology},
  volume={28},
  number={9},
  pages={3970--3977},
  year={2018},
  doi={10.1007/s00330-018-5351-3}
}

@article{koo_guideline_2016,
  title={A Guideline of Selecting and Reporting Intraclass Correlation Coefficients for Reliability Research},
  author={Koo, Terry K. and Li, Mae Y.},
  journal={Journal of Chiropractic Medicine},
  volume={15},
  number={2},
  pages={155--163},
  year={2016},
}

@book{krippendorff_content_2004,
  title={Content Analysis: An Introduction to Its Methodology},
  author={Krippendorff, Klaus},
  year={2004},
  publisher={Sage}
}

@article{landis_measurement_1977,
  title={The Measurement of Observer Agreement for Categorical Data},
  author={Landis, J. Richard and Koch, Gary G.},
  journal={Biometrics},
  volume={33},
  number={1},
  pages={159--174},
  year={1977},
  publisher={JSTOR},
}

@article{liao_evaluate_2019,
	title = {Evaluate the {Malignancy} of {Pulmonary} {Nodules} {Using} the 3-{D} {Deep} {Leaky} {Noisy}-{OR} {Network}},
	volume = {30},
	issn = {2162-2388},
	doi = {10.1109/TNNLS.2019.2892409},
	number = {11},
	journal = {IEEE transactions on neural networks and learning systems},
	author = {Liao, Fangzhou and Liang, Ming and Li, Zhe and Hu, Xiaolin and Song, Sen},
	year = {2019},
	pages = {3484--3495},
}

@article{mcgraw_forming_1996,
  title={Forming inferences about some intraclass correlation coefficients},
  author={McGraw, Kenneth O. and Wong, Seok P.},
  journal={Psychological Methods},
  volume={1},
  number={1},
  pages={30--46},
  year={1996}
}

@article{mikhael_sybil_2023,
	title = {Sybil: {A} {Validated} {Deep} {Learning} {Model} to {Predict} {Future} {Lung} {Cancer} {Risk} {From} a {Single} {Low}-{Dose} {Chest} {Computed} {Tomography}},
	volume = {41},
	issn = {1527-7755},
	doi = {10.1200/JCO.22.01345},
	number = {12},
	journal = {Journal of Clinical Oncology: Official Journal of the American Society of Clinical Oncology},
	author = {Mikhael, Peter G. and Wohlwend, Jeremy and Yala, Adam and Karstens, Ludvig and Xiang, Justin and Takigami, Angelo K. and Bourgouin, Patrick P. and Chan, PuiYee and Mrah, Sofiane and Amayri, Wael and Juan, Yu-Hsiang and Yang, Cheng-Ta and Wan, Yung-Liang and Lin, Gigin and Sequist, Lecia V. and Fintelmann, Florian J. and Barzilay, Regina},
	month = apr,
	year = {2023},
	pages = {2191--2200},
}

@article{nair_variable_2018,
	title = {Variable radiological lung nodule evaluation leads to divergent management recommendations},
	volume = {52},
	issn = {1399-3003},
	doi = {10.1183/13993003.01359-2018},
	number = {6},
	journal = {The European Respiratory Journal},
	author = {Nair, Arjun and Bartlett, Emily C. and Walsh, Simon L. F. and Wells, Athol U. and Navani, Neal and Hardavella, Georgia and Bhalla, Sanjeev and Calandriello, Lucio and Devaraj, Anand and Goo, Jin Mo and Klein, Jeffrey S. and MacMahon, Heber and Schaefer-Prokop, C. M. and Seo, Joon-Beom and Sverzellati, Nicola and Desai, Sujal R. and {Lung Nodule Evaluation Group} and {Lung Nodule Evaluation Group}},
	month = dec,
	year = {2018},
	pages = {1801359},
}

@article{national_lung_screening_trial_research_team_reduced_2011,
	title = {Reduced lung-cancer mortality with low-dose computed tomographic screening},
	volume = {365},
	issn = {1533-4406},
	doi = {10.1056/NEJMoa1102873},
	number = {5},
	journal = {The New England Journal of Medicine},
	author = {{National Lung Screening Trial Research Team} and Aberle, Denise R. and Adams, Amanda M. and Berg, Christine D. and Black, William C. and Clapp, Jonathan D. and Fagerstrom, Richard M. and Gareen, Ilana F. and Gatsonis, Constantine and Marcus, Pamela M. and Sicks, JoRean D.},
	year = {2011},
	pages = {395--409},
}

@misc{saab_capabilities_2024,
	title = {Capabilities of {Gemini} {Models} in {Medicine}},
	doi = {10.48550/arXiv.2404.18416},
	publisher = {arXiv},
    full_author = {Saab, Khaled and Tu, Tao and Weng, Wei-Hung and Tanno, Ryutaro and Stutz, David and Wulczyn, Ellery and Zhang, Fan and Strother, Tim and Park, Chunjong and Vedadi, Elahe and Chaves, Juanma Zambrano and Hu, Szu-Yeu and Schaekermann, Mike and Kamath, Aishwarya and Cheng, Yong and Barrett, David G. T. and Cheung, Cathy and Mustafa, Basil and Palepu, Anil and McDuff, Daniel and Hou, Le and Golany, Tomer and Liu, Luyang and Alayrac, Jean-baptiste and Houlsby, Neil and Tomasev, Nenad and Freyberg, Jan and Lau, Charles and Kemp, Jonas and Lai, Jeremy and Azizi, Shekoofeh and Kanada, Kimberly and Man, SiWai and Kulkarni, Kavita and Sun, Ruoxi and Shakeri, Siamak and He, Luheng and Caine, Ben and Webson, Albert and Latysheva, Natasha and Johnson, Melvin and Mansfield, Philip and Lu, Jian and Rivlin, Ehud and Anderson, Jesper and Green, Bradley and Wong, Renee and Krause, Jonathan and Shlens, Jonathon and Dominowska, Ewa and Eslami, S. M. Ali and Chou, Katherine and Cui, Claire and Vinyals, Oriol and Kavukcuoglu, Koray and Manyika, James and Dean, Jeff and Hassabis, Demis and Matias, Yossi and Webster, Dale and Barral, Joelle and Corrado, Greg and Semturs, Christopher and Mahdavi, S. Sara and Gottweis, Juraj and Karthikesalingam, Alan and Natarajan, Vivek},
	year = {2024},
}

@misc{sellergren_medgemma_2025,
	title = {{MedGemma} {Technical} {Report}},
	doi = {10.48550/arXiv.2507.05201},
	publisher = {arXiv},
    author = {Sellergren, Andrew and Kazemzadeh, Sahar and Jaroensri, Tiam and Kiraly, Atilla and Traverse, Madeleine and Kohlberger, Timo and Xu, Shawn and Jamil, Fayaz and Hughes, Cían and Lau, Charles and Chen, Justin and Mahvar, Fereshteh and Yatziv, Liron and Chen, Tiffany and Sterling, Bram and Baby, Stefanie Anna and Baby, Susanna Maria and Lai, Jeremy and Schmidgall, Samuel and Yang, Lu and Chen, Kejia and Bjornsson, Per and Reddy, Shashir and Brush, Ryan and Philbrick, Kenneth and Asiedu, Mercy and Mezerreg, Ines and Hu, Howard and Yang, Howard and Tiwari, Richa and Jansen, Sunny and Singh, Preeti and Liu, Yun and Azizi, Shekoofeh and Kamath, Aishwarya and Ferret, Johan and Pathak, Shreya and Vieillard, Nino and Merhej, Ramona and Perrin, Sarah and Matejovicova, Tatiana and Ramé, Alexandre and Riviere, Morgane and Rouillard, Louis and Mesnard, Thomas and Cideron, Geoffrey and Grill, Jean-bastien and Ramos, Sabela and Yvinec, Edouard and Casbon, Michelle and Buchatskaya, Elena and Alayrac, Jean-Baptiste and Lepikhin, Dmitry and Feinberg, Vlad and Borgeaud, Sebastian and Andreev, Alek and Hardin, Cassidy and Dadashi, Robert and Hussenot, Léonard and Joulin, Armand and Bachem, Olivier and Matias, Yossi and Chou, Katherine and Hassidim, Avinatan and Goel, Kavi and Farabet, Clement and Barral, Joelle and Warkentin, Tris and Shlens, Jonathon and Fleet, David and Cotruta, Victor and Sanseviero, Omar and Martins, Gus and Kirk, Phoebe and Rao, Anand and Shetty, Shravya and Steiner, David F. and Kirmizibayrak, Can and Pilgrim, Rory and Golden, Daniel and Yang, Lin},
	year = {2025},
}

@book{steyerberg_clinical_2019,
  title={Clinical Prediction Models},
  author={Steyerberg, Ewout W.},
  edition={2nd},
  year={2019},
  publisher={Springer}
}

@inproceedings{wang_medclip_2022,
  title={MedCLIP: Contrastive Learning from Unpaired Medical Images and Text},
  author={Wang, Zifeng and others},
  booktitle={EMNLP},
  year={2022}
}

\end{document}